\def\ARXIVVERSION{1}
\documentclass{article}

\ifdefined\ARXIVVERSION
\usepackage[preprint]{neurips_2026}
\else
\usepackage[dblblindworkshop]{neurips_2026}
\fi
\workshoptitle{Personalized, Aligned, Long-Term Memory for AI Systems (PALM)}

\usepackage[utf8]{inputenc}
\usepackage[T1]{fontenc}
\usepackage{hyperref}
\usepackage{url}
\usepackage{booktabs}
\usepackage{microtype}
\usepackage{xcolor}
\usepackage{graphicx}
\usepackage{float}
\hypersetup{colorlinks=true,citecolor=black,linkcolor=black,urlcolor=blue}

\title{Memory Depth and Reconstructed Context Width:\\
A Controlled Evaluation of Hierarchical Retrieval}

\ifdefined\ARXIVVERSION
\author{%
  Michael Andreev\thanks{Accepted at the PALM Workshop at NeurIPS 2026.} \\
  Independent Researcher \\
  \texttt{michael.andreev.ux@gmail.com}%
}
\else
\author{Anonymous Author(s)}
\fi

\begin{document}
\maketitle

\begin{abstract}
Long-term conversational memory is becoming an integral component of modern
LLM systems. Proposed architectures group records by topics and events,
construct hierarchies and graphs, and connect facts through causal and temporal
relations. We experimentally study the interaction between two memory
parameters: structural depth and the width of context supplied to the answer
model. Using EverMemBench, we evaluate depths D1--D4, core budgets of
1,024/2,048/4,096 tokens, and additional Production and Oracle conditions up to
the full archive. Increasing width from 1K to 4K improves Accuracy by
10.11--17.98 percentage points, whereas increasing depth provides no monotonic
gain. Beyond 8--16K, Production performance reaches a plateau while tokens per
correct answer continue to increase; Oracle preserves quality on full archives
of 68--71K tokens. These results motivate further investigation of large,
coherent context blocks instead of progressively deeper memory structures.
\end{abstract}

\section{Problem and research questions}

Hierarchical organization is used for interactive memory navigation and
temporal consolidation \citep{chen2023memwalker,li2026timem}, while recent
systems combine event graphs, multi-level representations, and controlled
retrieval \citep{wu2026gam,cao2026higmem,xu2026structmem}. Routing depth and
reconstructed context volume, however, usually vary as coupled properties of an
architecture rather than as jointly controlled parameters. A closely related
study finds that an additional progressive-disclosure level may fail to improve
quality and can reduce it, but focuses on document navigation rather than
long-term conversational memory \citep{he2026progressive}. We ask: \textbf{RQ1}
How do joint changes in Depth and Width affect answer quality and token
efficiency? \textbf{RQ2} Which measurable quantities identify a useful
Depth--Width operating point and inform memory architecture choices?

\section{System and experimental design}

The pipeline contains structured memory construction and context retrieval. A
Memory Writer converts dialogue into event cards linked to source messages. A
Hierarchy Writer assigns every card to a causal-thematic map without a preset
depth limit. Embeddings are built for nodes, cards, and dialogue chunks. At
answer time, a Production Router ranks nodes and cards by hybrid
lexical-semantic search. When cards are insufficient, a Context Packer retrieves
source chunks associated with selected nodes, fits them to the Width budget, and
restores chronological order. Depth is the number of accessible map levels;
shallower conditions remove complete lower levels and their cards.

EverMemBench contains long multi-party dialogues and question-answer-reference
triples \citep{hu2026evermembench}. The available data comprise 51,023 messages
(about 3.37M tokens) and 2,400 questions: 1,638 multiple-choice and 762
free-form, all with reference evidence. We selected three independent
topic-month episodes totaling 196,798 tokens, 2,996 messages, and 127 questions.
Dev is episode 02-2025-05 (38 questions); Main is episodes 04-2025-05 and
05-2025-05 (89 questions). Writer, map, Router, and evaluation procedures were
developed on Dev and frozen before Main. Questions, answers, and reference
evidence were excluded from memory construction and Production indexing.

\section{Dev protocol development}

The experiment began with a general $D\times W$ matrix and no restriction on
hierarchy depth. The Writers produced a naturally supported map reaching D4,
yielding D1--D4 conditions. Based on source context associated with map topics,
we selected incomplete-pool budgets of 1K/2K/4K. Accuracy increased with Width,
but Depth produced no monotonic gain: at 4K, Accuracy was
57.89/47.37/50.00/42.11\% for D1--D4. Evidence recall and Q-hit also increased,
but not proportionally to Accuracy.

Ambiguous core results motivated D1-wide (8K/16K/32K/full-map) and Oracle
(gold-only/4K/8K/16K/32K/full-archive) diagnostics. Oracle retained all gold
evidence while adding real dialogue context, separating memory failures from
the reader's use of wider inputs. It is not a strict Accuracy upper bound
because performance also depends on surrounding context and packing. Dev
suggested possible saturation, but one episode was insufficient for a
conclusion (Table~\ref{tab:dev}).

\begin{table}[t]
\caption{Dev Accuracy (\%) / TPCA (thousands of tokens).}
\label{tab:dev}
\centering
\scriptsize
\resizebox{\textwidth}{!}{%
\begin{tabular}{lrrrrrrrr}
\toprule
Retrieval & Gold & 1K & 2K & 4K & 8K & 16K & 32K & Full \\
\midrule
Production D1 & -- & 31.58/6.7 & 55.26/5.9 & 57.89/9.4 & 65.79/15.0 & 65.79/28.7 & 73.68/49.9 & 68.42/82.6 \\
Production D2 & -- & 26.32/7.9 & 42.11/7.7 & 47.37/11.4 & -- & -- & -- & -- \\
Production D3 & -- & 39.47/5.4 & 44.74/7.3 & 50.00/11.0 & -- & -- & -- & -- \\
Production D4 & -- & 28.95/7.3 & 36.84/9.0 & 42.11/12.5 & -- & -- & -- & -- \\
Oracle & 68.42/4.1 & -- & -- & 65.79/8.5 & 68.42/14.9 & 71.05/27.6 & 68.42/55.8 & 63.16/117.7 \\
\bottomrule
\end{tabular}}
\end{table}

Multiple-choice answers are scored deterministically. Two independent,
condition-blind judges (GPT-5.6 Sol and GPT-5.6 Terra) evaluate free-form
answers, with disagreements resolved by blinded manual adjudication. This
addresses known LLM-judge limitations such as position bias
\citep{zheng2023judge}. Metrics are Accuracy; tokens per correct answer (TPCA),
defined as total tokens divided by correct answers; evidence recall, the
fraction of gold source messages in packed context; and Q-hit, the fraction of
questions retrieving at least one gold source message. Accuracy and TPCA use
question-level bootstrap with 20,000 replicates \citep{koehn2004bootstrap}.
Paired comparisons use exact McNemar tests \citep{mcnemar1947} with Holm
correction. Before Main we froze data, maps, embeddings, the Grok 4.3 reader,
Router, 22 conditions, judging, statistics, and three hypotheses: \textbf{H1}
Width improves Accuracy within each Depth; \textbf{H2} Width-effect magnitude
depends on Depth; \textbf{H3} Width has a useful operating range, with narrow
contexts limiting quality and Accuracy plateauing after sufficient width while
TPCA rises.

\section{Frozen Main experiment}

The two frozen Main maps contained 489/512 cards and 90/108 nodes, reached D4,
and assigned every card to exactly one leaf. A post-construction evidence audit
gave 93.23/92.43\% coverage; gold annotations did not modify the maps. The 22
conditions comprised 12 Production combinations (D1--D4 $\times$ 1K/2K/4K),
four D1-wide conditions, and six Oracle conditions. We used 1,536-dimensional
\texttt{text-embedding-3-small} embeddings, hybrid score $\alpha=0.65$, scalable
branch/card caps, 320-token source chunks, and Grok 4.3 at temperature zero.
All 1,958 answers and required free-form judgments completed successfully.

\subsection{RQ1: Depth $\times$ Width}

\begin{table}[t]
\caption{Main Accuracy (\%) / TPCA (thousands of tokens), $n=89$.}
\label{tab:main}
\centering
\scriptsize
\resizebox{\textwidth}{!}{%
\begin{tabular}{lrrrrrrrr}
\toprule
Retrieval & Gold & 1K & 2K & 4K & 8K & 16K & 32K & Full \\
\midrule
Production D1 & -- & 42.70/4.1 & 53.93/5.3 & 60.67/8.4 & 64.04/14.9 & 69.66/26.7 & 67.42/54.3 & 69.66/89.1 \\
Production D2 & -- & 38.20/4.6 & 42.70/6.6 & 48.31/10.2 & -- & -- & -- & -- \\
Production D3 & -- & 39.33/4.6 & 48.31/6.0 & 55.06/9.3 & -- & -- & -- & -- \\
Production D4 & -- & 39.33/4.6 & 47.19/6.2 & 52.81/9.5 & -- & -- & -- & -- \\
Oracle & 69.66/3.8 & -- & -- & 66.29/7.8 & 68.54/14.3 & 67.42/28.8 & 73.03/52.4 & 74.16/108.1 \\
\bottomrule
\end{tabular}}
\end{table}

Width from 1K to 4K improved Accuracy at every depth: D1 42.70$\to$60.67\%,
D2 38.20$\to$48.31\%, D3 39.33$\to$55.06\%, and D4
39.33$\to$52.81\%. Effects were +10.11 to +17.98 points; all bootstrap 95\%
CIs excluded zero and Holm-adjusted McNemar tests gave $p\leq.0117$.
\textbf{H1 is supported.} Width effects varied by Depth, but no omnibus
interaction test was prespecified; \textbf{H2 was not formally tested}. Depth
gave no monotonic gain. At 4K the ordering was D1 60.67\%, D3 55.06\%, D4
52.81\%, D2 48.31\%. None of nine matched-width D1--Dx comparisons survived
joint Holm correction; the strongest was D1 versus D2 at 4K (+12.36 points,
$p_{\mathrm{Holm}}=.0665$).

\subsection{RQ2: Efficiency and sources of error}

Production D1 reached 64.04\% at 8K and 69.66\% at 16K, then did not improve:
32K yielded 67.42\% and full-map 69.66\%. TPCA rose from 14.9K to 26.7K,
54.3K, and 89.1K; wide comparisons beyond 8K were nonsignificant after Holm
correction. Together with the significant 1K--4K improvement, this supports
\textbf{H3}: the useful range lies between insufficiently narrow context and a
plateau around 8--16K.

Oracle scored 69.66\% on gold-only, 66.29--68.54\% at 4--16K, 73.03\% at 32K,
and 74.16\% on full-archive. No gold-only comparison was significant after Holm
correction ($p=1.0$). At 16K, Production D1 scored 69.66\% versus Oracle
67.42\%; a post-hoc paired test also found no difference ($p=.73$). Guaranteed
evidence is therefore not a strict Accuracy upper bound at every operating
point. Full archives of 68--71K tokens showed no reader degradation. The
Production--Oracle gap is compatible with writing and retrieval losses but does
not identify one causal mechanism.

\begin{figure}[H]
\centering
\includegraphics[width=\textwidth]{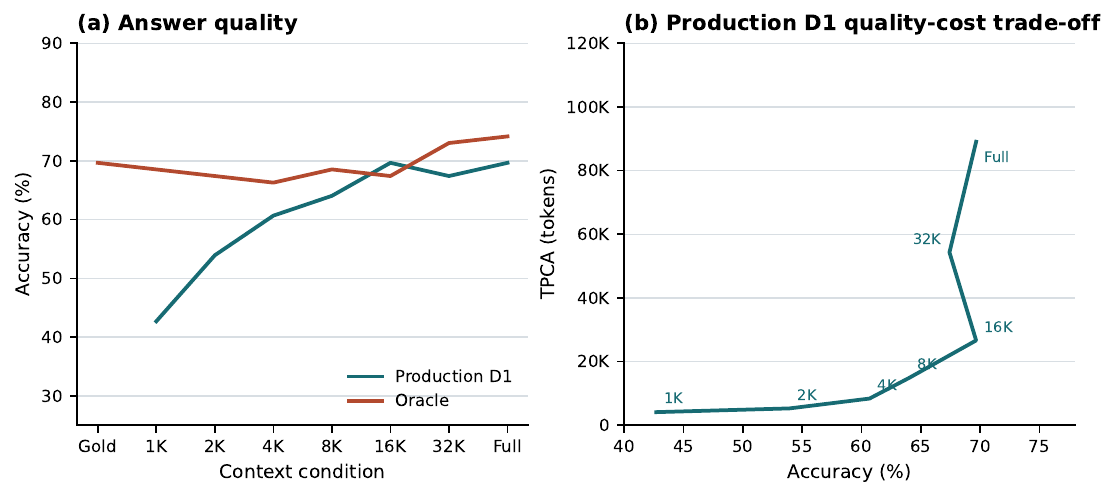}
\caption{Main Accuracy by context-width condition for Production D1 and Oracle
(left), and TPCA as a function of achieved Accuracy for Production D1 (right).}
\label{fig:main}
\end{figure}

At 4K, core recall was 39.27/32.69/37.58/38.54\% for D1--D4. Yet D1--4K Q-hit
of 96.63\% corresponded to only 60.67\% Accuracy. Retrieving some evidence does
not guarantee context sufficient for a correct answer.

\section{Discussion and conclusion}

Dev and Main differed beyond the core matrix. Dev Accuracy fell after a certain
width, whereas Main Production plateaued around 8--16K and Oracle preserved
quality through the full archive. The shifted saturation point is not a
universal model constant: it depends on system, data, and input construction
and requires deployment-specific calibration. This distinction between nominal
and effective context agrees with sensitivity to input position, length, and
task complexity reported elsewhere \citep{liu2024lost,hsieh2024ruler,paulsen2025mecw}.

Depth provided no monotonic improvement, but no D1--Dx difference survived
Holm correction; we therefore do not claim that depth is harmful. The narrower
result is that it provided no confirmed benefit here, while each added level
introduces further classification, linking, and routing decisions. Moreover,
the Context Packer consumed similar token volumes across D1--D4, so this design
does not test whether hierarchy can reduce token use.

Beyond 8--16K, TPCA grew disproportionately to Accuracy. Production D1 doubled
TPCA from 16K to 32K without improving quality; Oracle likewise approximately
doubled TPCA on its final step for little gain. Oracle gold-only and Production
Full both achieved 69.66\% Accuracy, but TPCA was 3.83K versus 89.07K: exact
oracle-guided selection required 23.2$\times$ fewer tokens per correct answer.
This illustrates precise retrieval's value but is not a causal estimate of all
Production losses.

The combined results motivate a hypothesis for future evaluation: memory may
work better as a shallow index into a small number of large, causally and
topically coherent blocks than as maximally atomic facts. Each block should
preserve a locally complete event and remain within the reader's effective
range; the index selects the block and the model reasons within it. This is
related to event-centric StructMem and two-level event-turn retrieval in HiGMem
\citep{xu2026structmem,cao2026higmem}, but follows here from controlled Depth
and Width variation. The unresolved problem is defining boundaries for events
spanning much of a dialogue. Duplication scales poorly, while one summary card
may lose coherence; cross-block links or a separate cross-cutting layer remain
untested directions.

\paragraph{Limitations.} One dataset, one reader, two Main maps, and depth only
to D4 limit generalization. Post-hoc, 21/89 questions were correct in all 22
conditions, 15 were never correct, and 53 were configuration-dependent. This
heterogeneity requires separate validation. The results do not establish a
universal block size or final architecture; they redirect study from maximum
decomposition depth toward coherent block size, boundary quality, and routing
cost at larger scales.

\bibliographystyle{plainnat}
\setlength{\bibsep}{2pt plus 0.3ex}

\ifdefined\ARXIVVERSION
\else
\newpage
\section*{NeurIPS Paper Checklist}

\begin{enumerate}

\item {\bf Claims}
    \item[] Question: Do the main claims made in the abstract and introduction accurately reflect the paper's contributions and scope?
    \item[] Answer: \answerYes{}
    \item[] Justification: The abstract and Sections 1 and 5 state the measured findings, scope, and limits without claiming a universal optimum.

\item {\bf Limitations}
    \item[] Question: Does the paper discuss the limitations of the work performed by the authors?
    \item[] Answer: \answerYes{}
    \item[] Justification: Section 5 identifies dataset, model, map-count, depth, question-heterogeneity, and generalization limitations.

\item {\bf Theory assumptions and proofs}
    \item[] Question: For each theoretical result, does the paper provide the full set of assumptions and a complete (and correct) proof?
    \item[] Answer: \answerNA{}
    \item[] Justification: The paper reports an empirical evaluation and contains no theoretical results or proofs.

    \item {\bf Experimental result reproducibility}
    \item[] Question: Does the paper fully disclose all the information needed to reproduce the main experimental results of the paper to the extent that it affects the main claims and/or conclusions of the paper (regardless of whether the code and data are provided or not)?
    \item[] Answer: \answerYes{}
    \item[] Justification: Sections 2--4 specify the data split, memory pipeline, conditions, models, retrieval limits, metrics, judging, and statistical procedures used for the main results.

\item {\bf Open access to data and code}
    \item[] Question: Does the paper provide open access to the data and code, with sufficient instructions to faithfully reproduce the main experimental results, as described in supplemental material?
    \item[] Answer: \answerNo{}
    \item[] Justification: The source dataset is externally distributed, and anonymized code is not linked during double-blind review. Code and reproducibility artifacts are planned for release upon acceptance.

\item {\bf Experimental setting/details}
    \item[] Question: Does the paper specify all the training and test details (e.g., data splits, hyperparameters, how they were chosen, type of optimizer) necessary to understand the results?
    \item[] Answer: \answerYes{}
    \item[] Justification: Sections 2--4 report the splits, controlled conditions, embedding model, Router parameters, reader settings, metrics, and evaluation protocol.

\item {\bf Experiment statistical significance}
    \item[] Question: Does the paper report error bars suitably and correctly defined or other appropriate information about the statistical significance of the experiments?
    \item[] Answer: \answerYes{}
    \item[] Justification: Sections 3 and 4 report question-level bootstrap confidence intervals, exact paired McNemar tests, Holm correction, replicate count, and random seed.

\item {\bf Experiments compute resources}
    \item[] Question: For each experiment, does the paper provide sufficient information on the computer resources (type of compute workers, memory, time of execution) needed to reproduce the experiments?
    \item[] Answer: \answerNo{}
    \item[] Justification: The paper reports token efficiency but does not provide local hardware specifications or complete wall-clock accounting because the principal model stages used hosted APIs.
    
\item {\bf Code of ethics}
    \item[] Question: Does the research conducted in the paper conform, in every respect, with the NeurIPS Code of Ethics \url{https://neurips.cc/public/EthicsGuidelines}?
    \item[] Answer: \answerYes{}
    \item[] Justification: The work uses an existing benchmark and hosted models for controlled evaluation and was conducted in accordance with the NeurIPS Code of Ethics.

\item {\bf Broader impacts}
    \item[] Question: Does the paper discuss both potential positive societal impacts and negative societal impacts of the work performed?
    \item[] Answer: \answerNo{}
    \item[] Justification: The study concerns memory-system evaluation and does not introduce a deployed system, but broader privacy and misuse implications of persistent memory are not analyzed in the paper.
    
\item {\bf Safeguards}
    \item[] Question: Does the paper describe safeguards that have been put in place for responsible release of data or models that have a high risk for misuse (e.g., pre-trained language models, image generators, or scraped datasets)?
    \item[] Answer: \answerNA{}
    \item[] Justification: The paper releases neither a trained model nor a newly collected dataset with high misuse risk.

\item {\bf Licenses for existing assets}
    \item[] Question: Are the creators or original owners of assets (e.g., code, data, models), used in the paper, properly credited and are the license and terms of use explicitly mentioned and properly respected?
    \item[] Answer: \answerNo{}
    \item[] Justification: EverMemBench and the hosted models are credited, but the final dataset version, canonical URL, and explicit license statement are still being verified before submission.

\item {\bf New assets}
    \item[] Question: Are new assets introduced in the paper well documented and is the documentation provided alongside the assets?
    \item[] Answer: \answerNA{}
    \item[] Justification: The paper does not introduce a new dataset or trained model; project code is not released as part of the anonymous submission.

\item {\bf Crowdsourcing and research with human subjects}
    \item[] Question: For crowdsourcing experiments and research with human subjects, does the paper include the full text of instructions given to participants and screenshots, if applicable, as well as details about compensation (if any)? 
    \item[] Answer: \answerNA{}
    \item[] Justification: The work does not involve crowdsourcing or research participants; manual adjudication was performed as part of the authors' evaluation workflow.

\item {\bf Institutional review board (IRB) approvals or equivalent for research with human subjects}
    \item[] Question: Does the paper describe potential risks incurred by study participants, whether such risks were disclosed to the subjects, and whether Institutional Review Board (IRB) approvals (or an equivalent approval/review based on the requirements of your country or institution) were obtained?
    \item[] Answer: \answerNA{}
    \item[] Justification: The work does not involve crowdsourcing or human-subject research.

\item {\bf Declaration of LLM usage}
    \item[] Question: Does the paper describe the usage of LLMs if it is an important, original, or non-standard component of the core methods in this research? Note that if the LLM is used only for writing, editing, or formatting purposes and does \emph{not} impact the core methodology, scientific rigor, or originality of the research, declaration is not required.
    \item[] Answer: \answerYes{}
    \item[] Justification: Sections 2--4 disclose the LLM-based Memory and Hierarchy Writers, Grok 4.3 reader, embedding model, and LLM judges used as core experimental components.

\end{enumerate}
\fi


{\small
\begin{thebibliography}{13}
\bibitem[Cao et~al.(2026)]{cao2026higmem} Cao, S., He, J., and Tan, F. HiGMem: A hierarchical and LLM-guided memory system for long-term conversational agents. \emph{Findings of ACL}, 2026.
\bibitem[Chen et~al.(2023)]{chen2023memwalker} Chen, H., Pasunuru, R., Weston, J., and Celikyilmaz, A. Walking down the memory maze: Beyond context limit through interactive reading. arXiv:2310.05029, 2023.
\bibitem[He et~al.(2026)]{he2026progressive} He, Y., Zhao, Y., Wang, J., and Chen, H. Is progressive disclosure all you need for long-context agents? arXiv:2607.17598, 2026.
\bibitem[Hsieh et~al.(2024)]{hsieh2024ruler} Hsieh, C.-P. et al. RULER: What's the real context size of your long-context language models? arXiv:2404.06654, 2024.
\bibitem[Hu et~al.(2026)]{hu2026evermembench} Hu, C. et al. Evaluating long-horizon memory for multi-party collaborative dialogues. \emph{KDD}, 2026. arXiv:2602.01313.
\bibitem[Koehn(2004)]{koehn2004bootstrap} Koehn, P. Statistical significance tests for machine translation evaluation. \emph{EMNLP}, 388--395, 2004.
\bibitem[Li et~al.(2026)]{li2026timem} Li, K. et al. TiMem: Temporal-hierarchical memory consolidation for long-horizon conversational agents. \emph{Findings of ACL}, 2026. arXiv:2601.02845.
\bibitem[Liu et~al.(2024)]{liu2024lost} Liu, N. F. et al. Lost in the middle: How language models use long contexts. \emph{TACL}, 2024.
\bibitem[McNemar(1947)]{mcnemar1947} McNemar, Q. Note on the sampling error of the difference between correlated proportions or percentages. \emph{Psychometrika}, 12:153--157, 1947.
\bibitem[Paulsen(2025)]{paulsen2025mecw} Paulsen, N. Context is what you need: The maximum effective context window for real world limits of LLMs. arXiv:2509.21361, 2025.
\bibitem[Wu et~al.(2026)]{wu2026gam} Wu, Z. et al. GAM: Hierarchical graph-based agentic memory for LLM agents. \emph{ACL}, 34647--34664, 2026.
\bibitem[Xu et~al.(2026)]{xu2026structmem} Xu, B. et al. StructMem: Structured memory for long-horizon behavior in LLMs. \emph{ACL}, 122--146, 2026.
\bibitem[Zheng et~al.(2023)]{zheng2023judge} Zheng, L. et al. Judging LLM-as-a-judge with MT-Bench and Chatbot Arena. arXiv:2306.05685, 2023.
\end{thebibliography}
}
\end{document}